\documentclass{article}

\usepackage{PRIMEarxiv}

\usepackage[utf8]{inputenc} % allow utf-8 input
\usepackage[T1]{fontenc}    % use 8-bit T1 fonts
\usepackage{hyperref}       % hyperlinks
\usepackage{url}            % simple URL typesetting
\usepackage{booktabs}       % professional-quality tables
\usepackage{amsfonts}       % blackboard math symbols
\usepackage{nicefrac}       % compact symbols for 1/2, etc.
\usepackage{microtype}      % microtypography
\usepackage{lipsum}
\usepackage{fancyhdr}       % header
\usepackage{graphicx}       % graphics
\graphicspath{{media/}}     % organize your images and other figures under media/ folder
\usepackage{amsmath}
\usepackage{float}
\floatstyle{plaintop}
\restylefloat{table}
\title{Segmentation Strategies in Deep Learning for Prostate Cancer Diagnosis: A Comparative Study of Mamba, SAM, and YOLO
}

\author{Ali Badiezadeh$^{a}$, Amin Malekmohammadi$^{b}$, Seyed Mostafa Mirhassani$^c$, Parisa Gifani$^d$, Majid Vafaeezadeh$^{e,}$\thanks{corresponding author}\\\\
\small$^{a}$ School of Electrical Engineering, Iran University of Science and Technology, Tehran, Iran\\
\small$^{b}$ School of Computer Engineering, Iran University of Science and Technology, Tehran, Iran\\
\small$^{c}$ School of Electrical Engineering, Shahrood Branch, Islamic Azad University, Shahrood, Iran\\
\small$^{d}$ Medical Sciences and Technologies Department, Science and Research Branch, Islamic Azad University, Tehran, Iran\\
\small $^{e}$ Biomedical Engineering Department, School of Electrical Engineering, Iran University of Science and Technology, Tehran, Iran\\
}

\begin{document}
\maketitle

\begin{abstract}
Accurate segmentation of prostate cancer histopathology images is crucial for diagnosis and treatment planning. This study presents a comparative analysis of three deep learning-based methods, Mamba, SAM, and YOLO, for segmenting prostate cancer histopathology images. We evaluated the performance of these models on two comprehensive datasets, Gleason 2019 and SICAPv2, using Dice score, precision, and recall metrics. Our results show that the High-order Vision Mamba UNet (H-vmunet) model outperforms the other two models, achieving the highest scores across all metrics on both datasets. The H-vmunet model's advanced architecture, which integrates high-order visual state spaces and 2D-selective-scan operations, enables efficient and sensitive lesion detection across different scales. Our study demonstrates the potential of the H-vmunet model for clinical applications and highlights the importance of robust validation and comparison of deep learning-based methods for medical image analysis. The findings of this study contribute to the development of accurate and reliable computer-aided diagnosis systems for prostate cancer. The code is available at http://github.com/alibdz/prostate-segmentation.
\end{abstract}

% keywords can be removed

\section{Introduction}
Prostate cancer, a widespread health concern impacting countless individuals globally, underscores the importance of accurate diagnostic methods to effectively gauge tumor aggressiveness and inform treatment plans. The Gleason scoring (grading) method, which has traditionally been, remains central to assessing the severity of prostate cancer. Nonetheless, this approach is fundamentally subjective and time-intensive, underscoring the necessity for cutting-edge techniques to boost diagnostic precision and efficiency. Recent strides in artificial intelligence (AI) and machine learning technologies present exciting opportunities for streamlining and enhancing the evaluation of histopathological images. Particularly, the utilization of image segmentation techniques within histopathology has become a critical focus of research, striving to distinguish between normal and malignant tissues with unmatched accuracy.  \\
Segmenting prostate's histopathology images poses unique challenges that require innovative solutions. One of the primary challenges is the variability and complexity of the tissue structures, which can lead to inconsistencies in segmentation outcomes. This variability is compounded by the fact that prostate's disease manifests differently across patients, necessitating a highly adaptable segmentation approach. Another significant challenge is the scale and detail of the images. High-resolution histopathology images often contain intricate details that are difficult to capture and analyze accurately. This complexity demands sophisticated algorithms capable of handling the vast amount of data while maintaining precision. \\
Moreover, traditional annotation processes involving such images are not only labor-intensive but also susceptible to human error, highlighting the need for automated solutions that can reliably annotate and segment the images. The lack of standardized protocols for Gleason grading further complicates the development of universally applicable segmentation algorithms, as there is a wide range of interpretations and practices among pathologists. \\
Lastly, the integration of AI and machine learning, notably those based on neural networks and deep learning models into clinical workflows presents its own set of challenges, including the need for extensive training data, the interpretability of model outputs, and the regulatory approval for clinical use.\\
Addressing these challenges requires a robust approach, combining advanced deep learning with rigorous validation against clinical standards. By overcoming these hurdles, we can enhance the accuracy and efficiency of prostate's disease diagnosis, ultimately improving patient outcomes.
In recent years, the deep learning methods of medical image analysis has witnessed significant advancements, particularly in the realm of histopathology image segmentation, grading, and classification. In \cite{cserbuanescu2020agreement} a two-stage deep learning system was developed for automated Gleason grading, achieving high accuracy with slide-level annotations only. Another study proposed an adaptive scoring mechanism that chooses among various annotations per image, optimizing Gleason grading performance and outperforming the STAPLE algorithm \cite{akhoondi2023semantic}. \\
Deep learning networks have also been used to classify prostate cancer images, with AlexNet showing substantial agreement with pathologists and perfect agreement with the majority vote, while GoogleNet had moderate agreement \cite{mun2021yet}. The Unsupervised Confidence Approximation (UCA) method was proposed to manage training with noisy data and make confident predictions concurrently, showing strong performance gains \cite{rabbani2023unsupervised}. \\
In addition, a novel domain adaptation approach was proposed to address class imbalance in the source domain and involve pseudo-labeled target data points in training, achieving competitive performance on public datasets for classifying prostate cancer images \cite{hosseini2023class}. In \cite{zhang2020gleason} a CNN model was used to segment TMA images into Gleason grade regions, achieving a high mean Dice score.
In \cite{foucart2023shortcomings} a review was conducted where investigated 21 digital pathology image segmentation challenges from 2010 to 2022, highlighting the importance of addressing inter-expert disagreement and the evaluation process chosen, as well as the quality control of various challenge elements to ensure accurate and reliable results. In another study, \cite{ciga2022self}, a contrastive learning method was used to pretrain a network on a large, diverse dataset of histopathology images, outperforming ImageNet pretraining on classification and regression tasks, and performing comparably for segmentation on smaller networks. The diversity of the pretraining dataset proved to be more important than the number of images. Furthermore, \cite{linmans2023latent} proposed the Latent Doctor Model (LDM), which is a stochastic classification framework utilizing a CVAE and DeepSet encoder, which can better model the full label distribution and capture annotator variability compared to baselines. The predictive uncertainty of the LDM correlates with inter-observer variability. \\
One notable approach is the probabilistic deep learning model proposed in \cite{schmidt2023probabilistic}, referred to as "Pionono". This method models inter- and intra-observer variability using probability distributions in latent space, and demonstrates superior performance over previous state-of-the-art approaches by a large margin. Additionally, Pionono provides probabilistic predictions that allow for uncertainty assessment and simulation of specific expert opinions. Pionono demonstrates improved predictive performance and estimated predictive uncertainty in prostate and breast cancer segmentation tasks, outperforming other models. Its ability to accurately model variability is highlighted as a significant advantage, especially in medical image analysis where subjective interpretations play a critical role. However, the model faces limitations in capturing detailed variations in annotator segmentations, prompting suggestions for increased model complexity to enhance its effectiveness. \\
Another approach that has shown promising results is the use of the STAPLE algorithm to synthesize different expert annotations, as described in \cite{qiu2022automatic}. This method was validated on the Vancouver Prostate Centre dataset and achieved good results in distinguishing between high-risk and low-risk cancer, as well as in distinguishing benign from malignant. Additionally, in \cite{mohsin2021automatic}, a UNET model with four different CNN architectures as the encoder was used. After preprocessing the data and using data augmentation, the UNET model with ResNet50 encoder achieved superior outcomes on both data sets. \\
In \cite{foucart2021processing} the impact of consensus methods and core-level scoring rules on the measured agreement between experts and the consensus in the analysis of the Gleason2019 dataset \cite{nir2018automatic} was investigated. The study found that the choice of these methods affected the measured agreement, which could influence the ranking of algorithms evaluated against the consensus. The authors used various consensus methods and scoring rules, and also employed a leave-one-out approach and MDS visualization to gain more insight into the relationships between experts and the consensus. Additionally, an advanced learning framework named MoMA has been introduced \cite{vuong2023moma}. It employs techniques such as knowledge distillation, momentum contrastive learning, and multi-head attention to facilitate the transfer of knowledge from a pre-trained teacher model to a student model that is trained with reduced dataset. The results show that MoMA is accurate and robust in transferring knowledge across varied applications, surpassing comparable techniques, and offering insights into effective learning strategy for a range of computational pathology contexts.\\
Over the years for segmentation and classification of medical and non-medical image modalities, a range of methods have been developed, primarily utilizing Convolutional Neural Networks \cite{ronneberger2015u} and Vision Transformers \cite{dosovitskiy2020image, vafaeezadeh2023carpnet,gifani2021automated,vafaeezadeh2021deep,vafaeezadeh2024ultrasound,gifani2023automatic,shalbaf2022automatic,vafaeezadeh2022automatic}. CNNs are renowned for their effectiveness in local feature extraction, essential for tasks requiring texture and fine detail analysis. However, CNNs struggle with capturing long-range dependencies due to their inherently local receptive field, even with larger convolution kernels. Additionally, processing high-resolution images with CNNs can be computationally expensive, demanding significant hardware resources.\\
In contrast, Vision Transformers (ViTs) are better in grasping global dependencies across an image, thanks to their self-attention mechanism. ViTs are scalable and can process larger images and more complex patterns effectively. Nonetheless, they require large amounts of memory, posing limitations in resource-constrained environments. The self-attention mechanism also adds to the model’s complexity, making ViTs harder to train and fine-tune.\\
State-space models (SSMs) have emerged as a promising alternative to both CNNs and ViTs, particularly with the introduction of the 2D-selective-scan (SS2D) method. SSMs efficiently capture long-range dependencies while maintaining linear complexity and facilitating parallel training, significantly reducing the time and computational resources required. However, SSMs’ focus on global receptive fields can introduce redundant information, not always relevant to the target features.\\
Recent advancements in state-space models, such as the Mamba algorithm, have incorporated time-varying parameters to enhance performance with fewer parameters than traditional Transformers. This led to the development of Vision Mamba (Vim), demonstrating superior hardware-aware design, saving substantial GPU memory while maintaining high performance in high-resolution vision tasks. Despite these advancements, existing models face challenges like excessive memory usage and the integration of redundant information. This highlights the need for models that balance the extraction of local and global features while minimizing computational costs and redundancy.\\
U-Mamba, a network for biomedical image segmentation that combines convolutional layers and State Space Sequence Models (SSMs) to handle long-range dependencies more effectively than traditional CNNs and Transformers \cite{ma2024u}.\\
Swin-UMamba model \cite{liu2024swin} is to enhance medical image segmentation by integrating multi-scale information and effectively modeling long-range global dependencies. It achieves this by combining the strengths of Mamba-based models in long sequence modeling with the advantages of ImageNet-based pretraining.\\
Building on advancements and addressing existing gaps, the High-order Vision Mamba UNet (H-vmunet) was developed \cite{wu2024h}, incorporating the strengths of Spectral-Spatial Models (SSMs), notably the 2D-selective-scan (SS2D) method, and extending it to high-order interactions to reduce redundant information while enhancing the global receptive field. The introduction of the Local-SS2D module significantly improves local feature extraction capabilities, making the model more adept at handling fine details essential in medical images. Furthermore, the H-vmunet model incorporates the High-order Visual State Space (H-VSS) module, enhancing feature extraction through multiple interaction orders. By integrating Channel and Spatial Attention modules in the skip connections, it further improves multi-level and multi-scale information fusion, accelerating convergence and increasing sensitivity to lesions of varying scales. Addressing the limitations of previous methods and introducing innovative modules, the H-vmunet aims to set a new standard in medical image segmentation, offering a more efficient, accurate, and scalable solution \cite{wu2024h}.\\
Among the deep learning methods, Mamba, SAM, and YOLO stand out for their potential in addressing the challenges of image segmentation in medical diagnostics. Mamba, a deep learning framework, is celebrated for its ability to handle large-scale image segmentation tasks efficiently. SAM, another cutting-edge method, specializes in semantic segmentation, providing detailed insights into the structural components of histopathology images. Meanwhile, YOLO is stands out for its ability to perform object detection accurately in real time, making it highly adaptable for the fast and accurate segmentation of complex structures in images.\\
This article delves into the comparative analysis of these three methods—Mamba, SAM, and YOLO—in the context of segmenting prostate cancer histopathology images. By exploring their strengths, limitations, and applicability in real-world clinical settings, we aim to shed light on the future directions of AI-assisted diagnostics in the field of oncology. The article is divided in such a way that at the beginning, we will have the method section which explores the Mamba, Yolo, and SAM frameworks within the context of image segmentation, starting with an explanation of their architectural designs. Following this we will explain the dataset used in this article. The Results section evaluates these frameworks' performance on various segmentation tasks, highlighting their strengths and limitations. The Discussion segment compares these results, analyzing each framework's unique attributes. Finally, the Conclusion summarizes our findings, emphasizing the significance of our observations for the future of image segmentation technologies.\\

\section{Materials}
Our study leverages two comprehensive datasets, namely the Gleason 2019 dataset \cite{nir2018automatic} and the SICAPv2 dataset \cite{silva2020going}, to explore the nuances of prostate cancer tissue imagery. Each dataset offers unique insights, yet they come with their own set of limitations that necessitate careful consideration during analysis.\\
As part of the author's study, it became clear that there exists five primary repositories for prostate cancer tissue imagery, with the Cancer Genome Atlas initiative holding the largest collection of approximately 720 biopsy slides \cite{weinstein2013cancer}. However, the lack of Gleason grade annotations at both slide and biopsy levels limits these datasets' applicability \cite{silva2020going}. In contrast, the database by Arvaniti et al.\cite{arvaniti2018automated} offers detailed pixel-level annotations for Gleason patterns across 886 smaller slide segments, though these do not fully capture the diversity of patterns in localized prostate cancer and benign conditions, hindering their utility for slide-level Gleason assessments. The PANDA challenge \cite{bulten2022artificial} introduced a substantial dataset, albeit with biopsy-level Gleason score annotations, not aligning with the research's goals. Additionally, two datasets, the Gleason19 challenge and SICAPv2, are thoroughly discussed as follows:
\subsection{Gleason dataset}
\begin{figure}[!h]
  \centering
    \includegraphics[width=0.8\columnwidth]{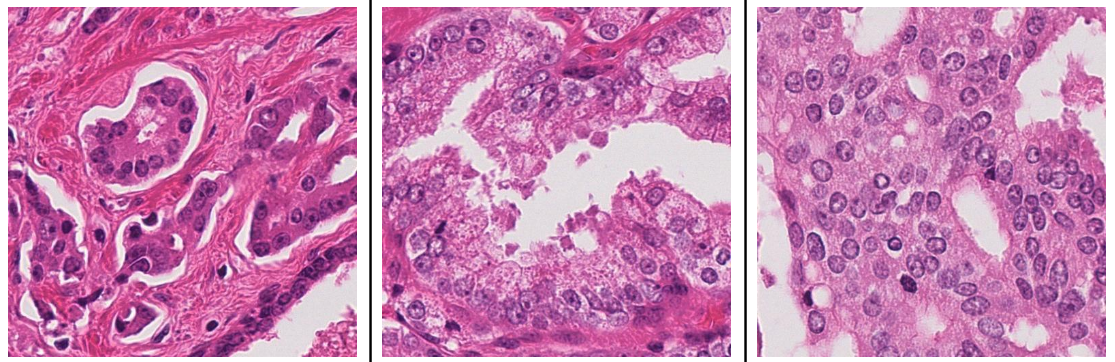}
    \caption{Examples of Gleason 2019 annotations. Left G3, middle G4, and right G5.}
  \label{fig:fig1}
\end{figure}

The 2019 Gleason challenge dataset, comprising 244 tissue microarray (TMA) images and their corresponding pixel-level annotations prepared by six pathologists, poses significant challenges in analyzing and making automated Gleason grading. One major challenge is the inter-observer variability in the annotations, which can affect the performance of automated grading models. This variability is evident in the fact that the annotations contain contradictory segmentations, making it difficult to select a single ground truth.\\
The limited number of Gleason images and their original resolution (5120 x 5120 pixels) make subsampling necessary. In particular, we extracted 512 x 512 patches with 50\% overlapping from each image and discarded the patches that lacked tissue texture. It is worth noting that for cross-validation, we followed a patient-based splitting strategy so that the patches belonging to a patient are confined to one fold. \\
Researchers have tackled the high inter-observer variability in prostate cancer grading, particularly in the Gleason 2019 Challenge dataset, by introducing innovative methods such as a dynamic scoring mechanism to balance annotations \cite{akhoondi2023semantic} and using Cohen's kappa coefficient to measure agreement among pathologists (mean inter-pathologist agreement of 0.42 ± 0.23) \cite{cserbuanescu2020agreement}. Other strategies include incorporating an "inter-expert prior" for medical image datasets \cite{rabbani2023unsupervised}, encoding annotations into separate channels \cite{zhang2020gleason}, modeling the full label distribution of experts \cite{linmans2023latent}, and synthesizing expert annotations with the STAPLE algorithm \cite{qiu2022automatic}. In our research, we employed max voting to aggregate expert annotations.\\
In addition, researchers have proposed various strategies to address imbalanced datasets. One approach involves using a weighted cross-entropy loss function during model training, adjusting class weights to balance class representation \cite{mun2021yet}. Another strategy combines data augmentation and resampling techniques, applying transformations like rotation and flipping to enhance dataset diversity and replicating images of underrepresented classes \cite{akhoondi2023semantic}. Some studies introduce novel algorithms, such as the focal loss function, which prioritizes minority class samples during training, and co-training approaches that leverage pseudo-labeled data \cite{hosseini2023class, zhang2020gleason}. Evaluation metrics like the macro F1 score are also utilized to ensure fairness across all classes, alongside techniques like random patch sampling from whole slide images to manage large datasets \cite{ciga2022self}. Lastly, significant improvements have been made by increasing dataset size through extensive data augmentation \cite{qiu2022automatic}. To counteract the class imbalance present in our dataset, where each class has a differing number of samples, we utilized data augmentation methods to achieve a balanced class representation. Specifically, the dataset underwent rotations up to 10 degrees and height shifts of up to 10\%, effectively increasing the number of instances for underrepresented classes. This augmentation strategy was crucial for preventing bias towards classes with larger sample sizes, ensuring a fair and balanced learning environment for the model. Figure \ref{fig:fig1} illustrates various examples of Gleason2019 annotations, showcasing different grades of severity.\\

\subsection{SICAPv2 dataset}
\begin{figure}[!h]
  \centering
    \includegraphics[width=0.8\columnwidth]{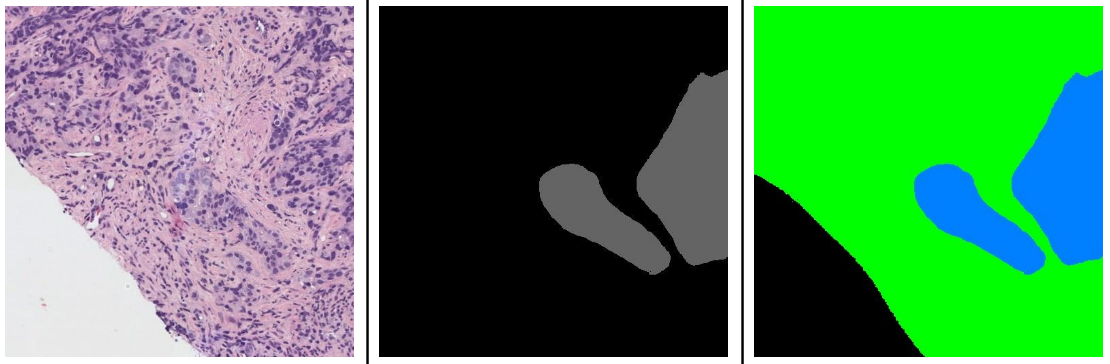}
    \caption{Segmentation map estimation on a sample image from SICAPv2 dataset. Left: original image, middle: provided segmentation map, right: processed segmentation map with Gleason grade (G4) annotation.}
  \label{fig:fig2}
\end{figure}

In the SICAPv2 dataset, pixel-level annotations are only provided for Gleason grades. To address this limitation, we implemented the technique described in \cite{anklin2021learning} to segregate these tissues from the surrounding background and create distinct masks for benign areas. Specifically, an image is first stain-normalized to reduce variability due to tissue preparation. The normalized image is then converted into a Tissue-Graph (TG), where nodes represent tissue regions as superpixels and edges represent interactions between these regions. The TG is created in three steps: constructing superpixels using the Simple Linear Iterative Clustering (SLIC) \cite{achanta2012slic} algorithm, extracting features from these superpixels, and defining the graph topology. Superpixels are merged hierarchically based on color similarity to form nodes, reducing complexity and enabling large image scaling. Each node is characterized by morphological and spatial features derived from image patches and superpixel centroids. The features are extracted with MobileNetV2 \cite{sandler2018mobilenetv2} pre-trained on ImageNet \cite{deng2009imagenet}. The TG's topology is defined by a Region Adjacency Graph (RAG)\cite{potjer1996region} that maps the spatial connectivity of superpixels. The result of this procedure on a sample image is summarized in figure \ref{fig:fig2}.

\section{Methods}
\subsection{Mamba}
Mamba emerges as a standout candidate, offering unique advantages that set it apart from other contemporary approaches. Mamba, a relatively new entrant in the domain of deep learning for image segmentation, has garnered attention for its ability to deliver superior performance in biomedical image segmentation tasks. Its core strength lies in combining the strengths of Convolutional Neural Networks (CNNs) and State Space Sequence Models (SSMs), effectively addressing long-range dependencies often encountered in medical imaging data. This combination enables Mamba to achieve impressive results in terms of Dice Similarity Coefficient (DSC) score across various medical imaging tasks, including the segmentation of 3D organ structures in CT and MRI scans, as well as precise delineation of cellular structures in microscopy images. \\
One of the key advantages of Mamba is its versatility and adaptability. It exhibits a self-configuring capability, allowing it to seamlessly adapt to various datasets. This feature enhances its scalability and flexibility, making Mamba a versatile tool in medical diagnostics and research. Furthermore, the integration of Mamba blocks within a U-Net architecture demonstrates the potential of State Space Models in biomedical image segmentation, setting a new benchmark in medical image segmentation.\\
Moreover, Mamba's robustness across varying input image sizes and its efficient processing of large-scale image datasets are notable advantages. These characteristics are crucial for practical applications where image resolutions can vary significantly, making Mamba computationally efficient and adaptable to real-world scenarios. Additionally, the leveraging of pretrained models in Mamba-based networks, such as Swin-UMamba and its variants, offers several benefits including enhanced segmentation accuracy, stable convergence, and mitigation of overfitting issues, data efficiency, and reduced computational resource consumption.\\
However, despite its numerous advantages, Mamba also faces certain challenges. The limited research and community support compared to more established methods like Transformers could pose potential hurdles in tackling complex tasks. Customization challenges arise due to the limited availability of pre-trained models, and adapting to Mamba may require additional learning efforts, especially for those familiar with alternative methodologies.\\
In conclusion, after careful consideration of the current landscape of deep learning methods for prostate cancer histopathology image segmentation, Mamba has been selected as the primary segmentation method for our study. Its superior performance, adaptability, and efficiency in handling large-scale datasets, coupled with the potential of integrating with existing architectures and the benefits of using pretrained models, make Mamba a compelling choice. As the field continues to evolve, Mamba stands poised to play a pivotal role in advancing the diagnosis and treatment of prostate cancer through accurate and efficient image segmentation. H-vmunet stands as one of the notable methodologies under the Mamba, we delve into with greater depth in subsequent section.\\

\subsection{H-vmunet method (Architecture)}
\begin{figure}[!h]
  \centering
    \includegraphics[width=0.85\columnwidth]{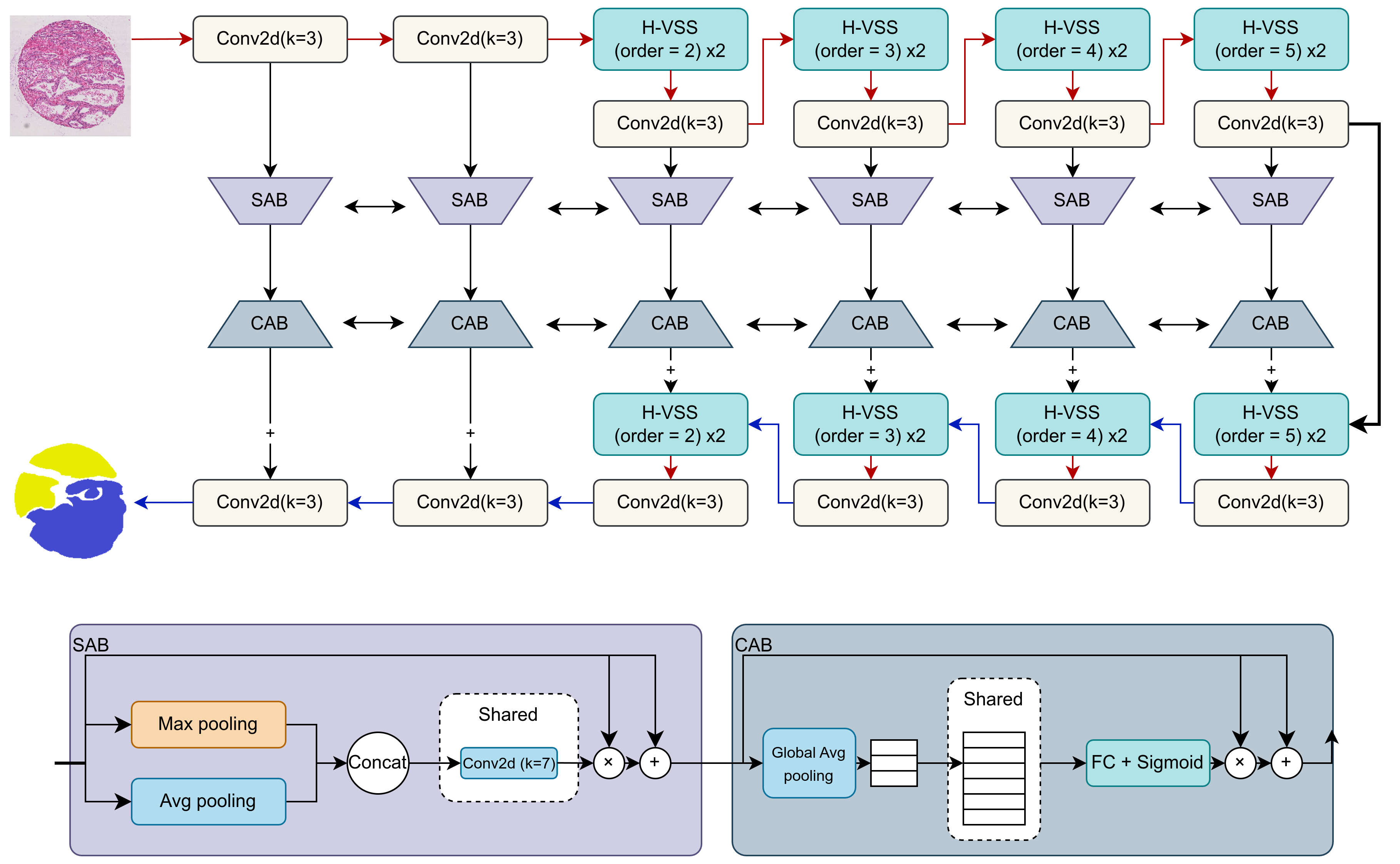}
    \caption{(Top) The H-vmunet model architecture overview. (Bottom) Multi-level and multi-scale information fusion module (SAB, CAB) \cite{wu2024h}.}
  \label{fig:fig3}
\end{figure}

The High-order Vision Mamba UNet (H-vmunet) is a model designed for medical image segmentation \cite{wu2024h}. It features a 6-layer U-shaped architecture which consists of an encoder, a decoder, and skip connections. These layers have channel sizes of 8, 16, 32, 64, 128, and 256. Layers 1 and 2 utilize standard convolution operations, while layers 3 to 6 integrate High-order Visual State Space (H-VSS) modules, each incorporating a standard convolution with a kernel size of 3. The H-VSS modules in these layers correspond to orders 2 through 5, enhancing feature extraction capabilities. The skip connections employ Channel Attention Bridge (CAB) and Spatial Attention Bridge (SAB) modules for multi-level and multi-scale information fusion \cite{ruan2022malunet}, improving the model's convergence speed and sensitivity to various lesion scales (Figure \ref{fig:fig3}).\\
The 2D-selective-scan technique is fundamental to the H-vmunet architecture which is proposed in \cite{zhu2024vision}. It involves a scan expansion process, an S6 block, and a scan merge operation, executed in four different scan order directions. Feature extraction is conducted along each direction using the S6 block \cite{gu2023mamba}, with the resulting sequences merged to restore the initial image size.\\
The High-order Visual State Space (H-VSS) module replaces the self-attention layer in traditional Transformer models with the High-order 2D-selective-scan (H-SS2D). This module includes LayerNorm (LN) layers, Multi-layer Perceptron (MLP), and H-SS2D operations, facilitating spatial interactions of arbitrary order. The H-SS2D is the core component for spatial feature extraction. The following equations represent the H-VSS expression:

\begin{equation}
y = \text{HS}[\text{LN}(x)] + x
\end{equation}

\begin{equation}
\text{Out} = \text{MLP}[\text{LN}(y)] + y
\end{equation}

Where LN is the LayerNorm layer, HS is the H-SS2D operation, and MLP is the multilayer perceptron \cite{wu2024h}.

\begin{figure}[!h]
  \centering
    \includegraphics[width=1.0\columnwidth]{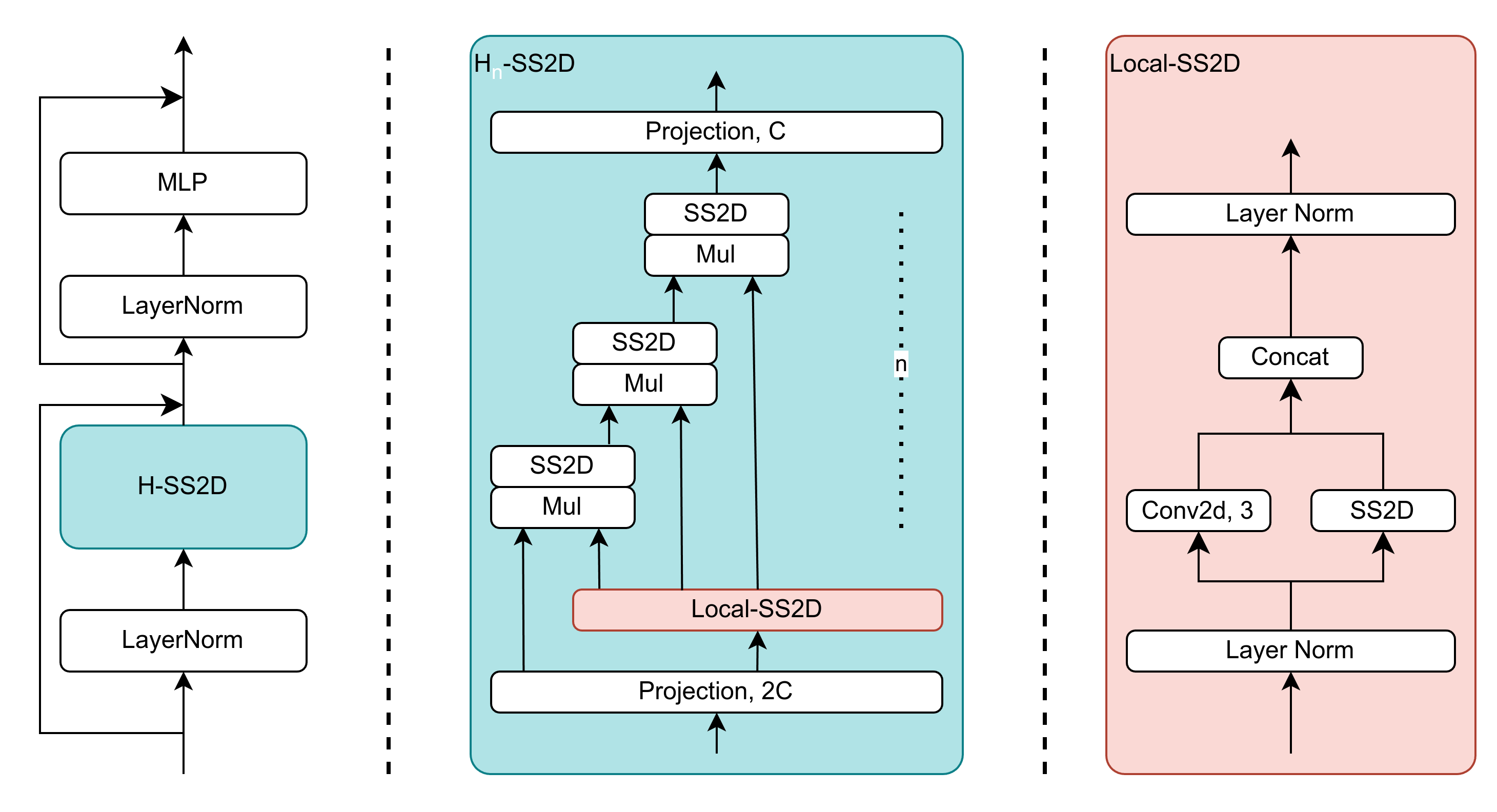}
    \caption{(Left) Architecture of the H-VSS module. (Middle) Overview of n-order 2D-selective-scan module. (Right) Architecture of Local-SS2D module \cite{wu2024h}}
  \label{fig:fig4}
\end{figure}

The 1-order 2D-selective-scan operation, referred to as H1-SS2D, is a fundamental component of the H-vmunet architecture. It begins by transforming the input feature map in four different directions: horizontal, vertical, and two diagonal directions. Each direction undergoes an independent 2D-selective-scan, which extracts relevant features from the input. To enhance feature extraction, the Local-SS2D module combines local convolutional operations with the global 2D-selective-scan. This dual approach allows the model to effectively balance local detail and global context, leading to a comprehensive feature representation \cite{wu2024h}. The higher-order 2D-selective-scan generalizes this approach to any order (Figure \ref{fig:fig4}). For higher-order scans, the process recursively applies the previous order scan. This hierarchical scanning captures increasingly complex spatial dependencies, building on the earlier scans. Each higher-order scan incorporates the findings of the previous scans, allowing the model to extract features at multiple scales and hierarchies. This recursive, layered approach enables the model to capture both fine-grained details and broad contextual information effectively \cite{wu2024h}.\\

\subsection{SAM}
\begin{figure}[!h]
  \centering
    \includegraphics[width=1.0\columnwidth]{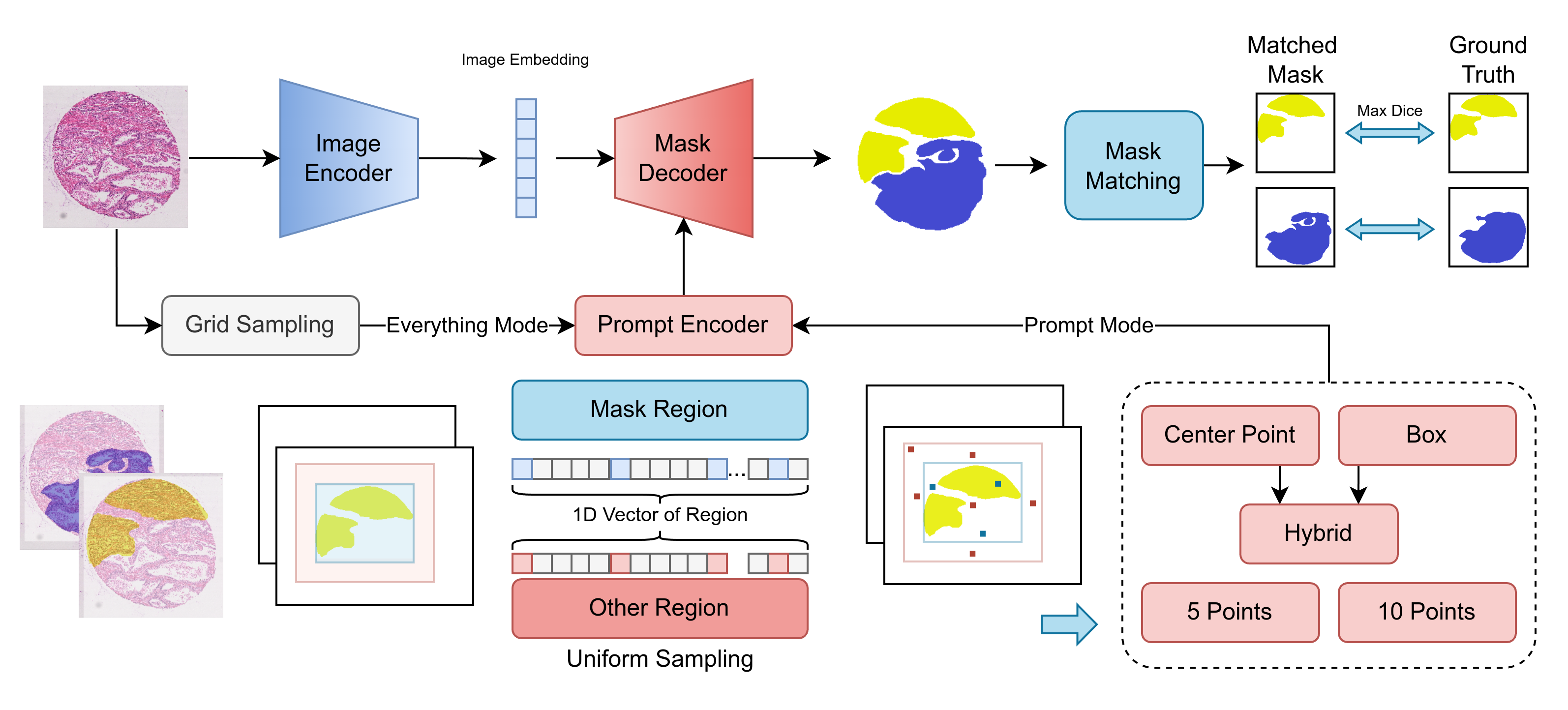}
    \caption{The pipeline of SAM in \cite{huang2024segment}}
  \label{fig:fig5}
\end{figure}

SAM sets itself apart from conventional segmentation frameworks by introducing a novel promptable segmentation task, underpinned by a flexible model architecture capable of processing prompts and a vast, diverse array of training data. A data engine is employed to create a cyclical process where the model aids in data collection, which in turn is used to improve the model's performance \cite{huang2024segment}. \\
SAM's architecture is meticulously designed to address the complexities of image segmentation tasks, featuring three interconnected components: an image encoder, a prompt encoder, and a mask decoder (see Figure \ref{fig:fig5}). The image encoder, built on the ViT backbone, was pretrained using the masked autoencoder (MAE) technique \cite{he2022masked}. It processes the input image, generating a rich embedding that captures the spatial and semantic features of the scene. This component plays a crucial role in SAM's ability to understand the context and details of the image, facilitating accurate segmentation. The Prompt Encoder is designed to interpret human-provided prompts, whether they are textual descriptions, points, or bounding boxes indicating the areas of interest. This capability enables SAM to perform zero-shot segmentation, adapting to new tasks without explicit retraining, thereby enhancing its versatility and applicability. The Mask Decoder, a lightweight yet effective component, takes the embeddings produced by the encoders and generates segmentation masks. This process involves predicting pixel-wise labels that delineate the objects of interest within the image, achieving precise localization and segmentation. \\
To effectively utilize SAM for specific tasks, fine-tuning is essential. This involves adjusting the model's hyperparameters, such as learning rates and batch sizes, to optimize its performance on the target dataset. Careful selection and tuning of these parameters are critical to avoid overfitting and ensure that SAM adapts well to new data, maintaining its high level of accuracy and efficiency. Moreover, SAM's architecture benefits from a vast dataset, comprising over 1 billion masks derived from 11 million images. This extensive training data equips SAM with robust generalization capabilities, enabling it to excel in a wide range of segmentation tasks with minimal adjustments. \\
SAM offers both automatic and manual modes during testing. In the automatic mode, users simply input an image, and SAM automatically generates all masks. In the manual mode, users supply additional inputs such as boxes, points, masks, and text to provide SAM with more detailed information about the objects to be segmented \cite{huang2024segment}. 

Figure 5: The pipeline of SAM in \cite{huang2024segment}

\subsection{YOLO}
YOLO (You Only Look Once) is a revolutionary object detection algorithm that has significantly impacted the field of computer vision by offering real-time object detection capabilities. Initially introduced by Joseph Redmon, YOLO has evolved through several iterations, with YOLOv8 representing the latest advancement in this series. A review study highlighted several key advantages of YOLO (You Only Look Once) in medical image analysis:
\begin{enumerate}
    \item Real-Time Performance: YOLO is noted for its real-time object detection capabilities, which are essential for urgent medical interventions and clinical decisions. This real-time functionality facilitates swift image processing, contributing to timely disease identification, therapeutic strategy formulation, and disease evolution tracking.
    \item Versatility and Generalizability: YOLO's flexibility permits it to identify and locate a wide range of anatomical elements, abnormalities, growths, and other pertinent medical items across various medical imaging modalities. Its adaptability ensures reliable performance even in new contexts or unforeseen scenarios, a significant advantage in the ever-evolving medical landscape.
    \item High Accuracy: The paper mentions that YOLO outperforms other existing methods in various medical object detection tasks, including the identification of lesions, categorization of skin lesions, detection of retinal anomalies, cardiac abnormalities and brain tumor delineation This precision stems from its capacity to grasp generalized object representations, exceeding benchmarks set by models such as the Deformable Parts Model (DPM) \cite{srikar2022improved} and Faster RCNN \cite{drid2020object}.
    \item Efficiency: While maintaining high precision, YOLO ensures operational efficiency, vital for real-world healthcare implementations. Its design supports quicker object detection compared to conventional two-phase methodologies, enhancing its suitability for immediate use cases \cite{patel2023comprehensive}.
    \item Adaptability to Different Medical Domains: YOLO demonstrates efficacy across a spectrum of medical disciplines, encompassing medical imaging, surgical processes, and personal protective equipment detection. This versatility emphasizes its extensive role in healthcare solutions \cite{qureshi2023comprehensive}.
\end{enumerate}

These attributes position YOLO as a formidable asset for medical image analysis, empowering medical practitioners to diagnose and manage a variety of health issues with proficiency and speed. YOLOv8, developed by the Ultralytics team, builds upon the foundational principles of YOLO but incorporates major improvements in speed, accuracy, and efficiency, making it a state-of-the-art system for object detection tasks \cite{ravsic2023detection, yu2023pharmacy, liu2024socr}. YOLOv8, in particular, offers several key features \cite{sohan2024review} that are advantageous for medical image analysis:
\begin{itemize}
    \item Multi-Scaled Detection: YOLOv8 excels in detecting objects of various sizes within an image, which is essential for medical imaging where objects can vary greatly in scale \cite{farooq2024improved}.
    \item Improved Accuracy: Despite its speed, YOLOv8 maintains high accuracy in object detection, making it reliable for medical applications where precision is paramount.
    \item Improved Accuracy: Despite its speed, YOLOv8 maintains high accuracy in object detection, making it reliable for medical applications where precision is paramount.
    \item Flexible Framework: YOLOv8's unified architecture seamlessly integrates object detection, instance segmentation, and image classification capabilities, enabling a comprehensive approach to medical image analysis. This versatility allows for the localization, delineation, and categorization of anatomical structures, pathological regions, and other clinically relevant features within medical imagery.
    \item Pre-Trained Models: Offering pre-trained models for different tasks, YOLOv8 can be easily adapted for specific medical imaging tasks, saving time and resources.
    \item Advanced Customization Options: YOLOv8 provides advanced customization options, enabling tailored solutions for unique medical imaging challenges.
    \item Anchor-Free Architecture: Unlike previous versions that relied on predefined anchor boxes, YOLOv8 employs an anchor-free architecture. This innovation allows for direct prediction of object centers without relying on fixed anchor shapes, enhancing flexibility and accuracy in object detection
    \item Advanced Backbone Network: YOLOv8 introduces a new backbone network, contributing to its improved performance. This advanced architecture supports more efficient feature extraction, further boosting the model's overall detection capabilities.
\end{itemize}

These features make YOLOv8 ones of excellent choices for our study, as they address the critical needs of medical image analysis, including the ability to quickly and accurately detect and classify objects within medical images, thereby facilitating more informed diagnostic decisions and treatment plans. In this study we used YOLOv8m (medium) model architecture, offering a compelling trade-off between inference speed and accuracy for applications.

\section{Results}
We use various segmentation metrics to quantitatively assess different methods. The Dice coefficient remains unaffected by changes in pattern size because it assesses the proportion of overlap between two sets rather than their absolute sizes. Consequently, it delivers consistent similarity scores as long as the overlap proportion is unchanged, regardless of the size of the segmented regions. Additionally, since segmentation can be viewed as pixel-level classification, we also utilize precision and recall, defined as follows:

\begin{equation}
\text{Precision} = \frac{\text{TP}}{\text{TP} + \text{FP}}
\end{equation}

\begin{equation}
\text{Recall} = \frac{\text{TP}}{\text{TP} + \text{FN}}
\end{equation}

\begin{equation}
\text{F-Score} = \frac{2 \times (\text{Precision} \times \text{Recall})}{\text{Precision} + \text{Recall}}
\end{equation}

\begin{equation}
\text{Accuracy} = \frac{\text{TP} + \text{TN}}{\text{TP} + \text{TN} + \text{FP} + \text{FN}}
\end{equation}

With TP, FP, TN, and FN indiciating true positives, false positives, true negatives, and false negatives, respectively.
These metrics highlight the challenge in differentiating between four classes in Gleason2019 and SICAPv2 dataset. It is worth mentioning that due to severe class imbalance in these datasets, we opt for a weighted average over class metrics to provide a more comprehensive measurement of overall performance. This approach ensures that the performance metrics reflect the proportion of each class in the dataset, rather than being skewed by the performance on less frequent Gleason patterns.
We compared the YoloS, SAM, and H-vmunet models on the Prostate2019 and SICAP-V2 datasets using Dice score, precision, and recall metrics, as presented in Tables \ref{tabel:gl} and \ref{tabel:si}. On the Prostate2019 dataset, H-vmunet outperformed the other models, achieving the a dice score of 0.92 (highest score) across all metrics, which highlights its effectiveness in segmenting Gleason images. This superior performance can be attributed to the advanced architecture of H-vmunet, which leverages vision mamba to better capture and differentiate complex tissue structures. Similarly, on the SICAP-V2 dataset, H-vmunet again achieved the best results (dice score of 0.68) compared to SAM and YoloS, demonstrating its robustness and generalization capability across different types of pathological images. The consistent high performance of H-vmunet suggests that its sophisticated model design, which includes enhanced feature extraction and segmentation strategies, is particularly well-suited for medical image analysis, making it more effective in handling the variability and complexity inherent in these datasets.
The H-vmunet showcases notable performance metrics, with Precision and Recall values both standing at 0.68 for the SICAP-V2 and 0.92. This uniformity in scores indicates a balanced approach in identifying true positives and minimizing false negatives, suggesting a well-rounded performance of the H-vmunet method in the evaluated dataset. Such consistency across these key metrics underscores the method's reliability and effectiveness in classification tasks, offering promising insights for further application and refinement in related domains.

\begin{table}[h!]
\centering
\caption{Comparison results on the prostate 2019 dataset }
\begin{tabular}{lcccc}
\toprule
 Method & DSC (F1) & Precision & Recall \\
\midrule
YoloS(avg)&              0.83 ± 0.015 & 0.84 ± 0.014 & 0.83 ± 0.016  \\
SAM (avg)  &             0.86 ± 0.120 & 0.87 ± 0.100 & 0.85 ± 0.110 \\
H-vmunet (avg)&          0.92 ± 0.062 & 0.92 ± 0.061 & 0.92 ± 0.064  \\
\bottomrule
\end{tabular}
\label{tabel:gl}
\end{table}

\begin{table}[h!]
\centering
\caption{Comparison results on the SICAP-V2 dataset }
\begin{tabular}{lcccc}
\toprule
 Method & DSC (F1) & Precision & Recall \\
\midrule
YoloS(avg)&              0.65 ± 0.025 & 0.64 ± 0.027 & 0.65 ± 0.024  \\
SAM (avg)  &             0.66 ± 0.12 & 0.67 ± 0.049 & 0.66 ± 0.027 \\
H-vmunet (avg)&          0.68 ± 0.034 & 0.68 ± 0.047 & 0.68 ± 0.031  \\
\bottomrule
\end{tabular}
\label{tabel:si}
\end{table}

\section{Discussion}
The comparative analysis presented in this study underscores the significant role of deep learning models in advancing prostate cancer diagnosis through histopathology image segmentation. The comparative analysis of Mamba, SAM, and YOLO models for segmenting prostate cancer histopathology images reveals significant insights into the capabilities and limitations of each method. The High-order Vision Mamba UNet (H-vmunet) model emerged as the most effective across both the Gleason 2019 and SICAPv2 datasets, showcasing its robustness and adaptability in handling complex medical imaging tasks. This superior performance underscores the importance of innovative architectural designs and advanced feature extraction techniques in achieving accurate segmentation outcomes. \\
H-vmunet's standout performance can be attributed to several key factors. Firstly, its integration of High-order Visual State Space (H-VSS) modules and 2D-selective-scan operations enables efficient and sensitive lesion detection across varying scales. This capability is crucial for medical image analysis, where the ability to discern subtle differences between normal and malignant tissues is paramount, especially when it comes to accurately identifying and delineating malignant tissues within the intricate and varied structures found in prostate histopathology images. Secondly, The model's architecture not only facilitates multi-level and multi-scale information fusion, thereby enhancing its sensitivity to lesions of varying sizes and improving convergence speed, but also balances local detail extraction with global context understanding through the Local-SS2D module and Channel and Spatial Attention modules in skip connections, further accelerating convergence and increasing sensitivity to lesions across different scales. The architectural innovations significantly contribute to H-vmunet's superior segmentation accuracy compared to SAM and YOLO models, collectively enhancing the model's Dice score, precision, and recall metrics observed in our study.\\
Despite the impressive performance of H-vmunet, it is important to acknowledge potential challenges and limitations. One such challenge is the computational complexity associated with high-order visual state spaces and 2D-selective-scan operations. While these techniques enhance segmentation accuracy, they may also increase processing time and resource requirements, potentially limiting real-time applicability in clinical settings. Additionally, the effectiveness of H-vmunet, like any deep learning model, heavily relies on the quality and diversity of training data. Insufficient training data or biases within the dataset could impact the model's generalization capabilities, affecting its performance on unseen images.\\
SAM and YOLO, while demonstrating promising results in various segmentation tasks, did not outperform H-vmunet in our study. SAM's strength lies in its flexibility and adaptability, particularly its ability to perform zero-shot segmentation, which is crucial in various applications. However, its reliance on human-provided prompts may limit its autonomy in clinical settings where immediate and accurate segmentation is required without manual intervention. Despite its innovative promptable segmentation task and flexible architecture capable of processing prompts, SAM did not match the segmentation accuracy of H-vmunet on both datasets. This discrepancy could be due to SAM's dependence on human-provided prompts and its architectural design focusing on generalizability across various image types, potentially not fully exploiting the specific characteristics of prostate cancer histopathology images.\\
YOLO, renowned for its real-time object detection capabilities, provides substantial benefits in terms of speed and efficiency. However, its performance in segmenting prostate cancer histopathology images did not reach the level of H-vmunet, likely because of the intricate and variable nature of medical images necessitating advanced feature extraction techniques that go beyond simple object detection. Furthermore, even though YOLOv8 showed enhancements in speed, accuracy, and efficiency, it still lagged behind H-vmunet in histopathology image segmentation. This indicates that while YOLO shines in object detection tasks, the detailed requirements of histopathology image segmentation—demanding precise delineation of tissue structures—are better served by the specialized architectures and methodologies utilized by H-vmunet.\\
The findings of this study hold significant implications for clinical practice, highlighting the critical importance of accurate segmentation of prostate cancer histopathology images for diagnosis, treatment planning, and ultimately, patient outcomes. The demonstrated effectiveness of H-vmunet suggests its potential as a highly valuable tool within clinical workflows, offering enhanced precision in identifying malignant tissues. However, the successful integration of such models into clinical settings necessitates overcoming computational challenges and ensuring rigorous validation against established clinical standards. Additionally, the continuous evolution and adaptation of models like H-vmunet are imperative to maintain alignment with advancements in medical imaging technologies and the ever-changing landscape of diagnostic needs. While the model's robust performance across various datasets indicates its broad applicability in clinical environments, making it a versatile asset for oncologists and pathologists, it is crucial to acknowledge the challenges involved in integrating AI-assisted diagnostic tools into clinical workflows. These include the necessity for extensive training data, the interpretability of model outputs, and obtaining regulatory approval for clinical use, alongside the commitment to ongoing research and validation against clinical standards to guarantee the reliability and safety of these tools.\\
Future research directions for H-vmunet should focus on enhancing its computational efficiency to support real-time applications without sacrificing accuracy. Exploring strategies to boost model generalization, such as incorporating more diverse training datasets or employing data augmentation techniques, could further enhance performance on unseen images. Additionally, integrating H-vmunet with other AI-driven diagnostic tools could offer comprehensive solutions for prostate cancer diagnosis, potentially transforming clinical decision-making processes. Looking ahead, future research should concentrate on refining and extending the capabilities of the H-vmunet model. This might involve exploring architectural enhancements, such as incorporating attention mechanisms or other advanced techniques, to improve segmentation accuracy. Investigating the model's performance on other types of medical images and diseases could broaden its applicability and impact on healthcare.\\
This comparative study emphasizes the potential of advanced deep learning models, notably the High-order Vision Mamba UNet (H-vmunet), in significantly enhancing the accuracy and efficiency of prostate cancer diagnosis through improved histopathology image segmentation. As artificial intelligence continues to advance, ongoing research and development efforts are pivotal in unlocking its full potential to drive progress in patient care and outcomes.

\section{Conclusion}
This study presents a comprehensive comparative analysis of three deep learning-based methods, Mamba, SAM, and YOLO, for segmenting prostate cancer histopathology images, utilizing the Gleason 2019 and SICAPv2 datasets. The findings reveal that the High-order Vision Mamba UNet (H-vmunet) model surpasses the other models in terms of segmentation accuracy, achieving the highest scores across all evaluated metrics. This superiority is attributed to H-vmunet's innovative architecture, which integrates high-order visual state spaces and 2D-selective-scan operations, enabling efficient and sensitive lesion detection across different scales.\\
The results underscore the potential of H-vmunet for clinical applications, particularly in enhancing the accuracy and efficiency of prostate cancer diagnosis through histopathology image segmentation. They highlight the importance of robust validation and comparison of deep learning-based methods for medical image analysis, contributing to the development of accurate and reliable computer-aided diagnosis systems for prostate cancer.\\
While Mamba, SAM, and YOLO demonstrate promising aspects, such as Mamba's adaptability and SAM's flexibility in processing prompts, their performance in segmenting prostate cancer histopathology images does not match that of H-vmunet. This suggests that for the specific task of prostate cancer histopathology image segmentation, specialized architectures and methodologies, like those employed by H-vmunet, are more effective.\\
The integration of such advanced models into clinical workflows necessitates addressing computational challenges and ensuring rigorous validation against established clinical standards. Future research should focus on enhancing H-vmunet's computational efficiency, boosting model generalization, and investigating its application in other medical imaging tasks. This study contributes to the broader discussion on the future of AI-assisted diagnostics in oncology, emphasizing the critical role of advanced deep learning models in improving patient outcomes through precise and efficient image segmentation.

% \section*{Acknowledgments}
% This was was supported in part by......

%Bibliography
\bibliographystyle{unsrt}  
\bibliography{references}

\end{document}